\documentclass[letters,10 pt, journal, twoside]{IEEEtran}
\ifCLASSINFOpdf
\else
\fi

\usepackage{graphicx} % for pdf, bitmapped graphics files
\usepackage{epsfig} % for postscript graphics files
\usepackage{mathptmx} % assumes new font selection scheme installed
\usepackage{times} % assumes new font selection scheme installed
\usepackage{amsmath} % assumes amsmath package installed
\usepackage{amssymb}  % assumes amsmath package installed
\usepackage{multirow}
\usepackage{mathtools, cases}
\usepackage{soul}
\usepackage{soulutf8}
\usepackage{dsfont}
\usepackage{tikz}
\usepackage{wrapfig}
\usepackage{color}
\usepackage{xcolor}
\usepackage{hyperref}
\usepackage{caption, subcaption, multicol}

\newcommand{\method}{\textsc{MASK}}

\begin{document}
%
% paper title
% Titles are generally capitalized except for words such as a, an, and, as,
% at, but, by, for, in, nor, of, on, or, the, to and up, which are usually
% not capitalized unless they are the first or last word of the title.
% Linebreaks \\ can be used within to get better formatting as desired.
% Do not put math or special symbols in the title.
\title{Towards Embedding Dynamic Personas in Interactive Robots: Masquerading Animated Social Kinematic (MASK)}

\author{Jeongeun Park$^{1}$, Taemoon Jeong$^{1}$, Hyeonseong Kim$^{1}$, Taehyun Byun$^{1}$, Seungyoun Shin$^{1}$,\\ Keunjun Choi$^{2}$, 
Jaewoon Kwon$^{3}$, Taeyoon Lee$^{3}$, Matthew Pan$^{4}$, and Sungjoon Choi$^{1}$ %
\thanks{Manuscript received: March 14, 2024; Revised June 25, 2024; Accepted July 22, 2024.}%Use only for final RAL version
\thanks{This paper was recommended for publication by Associate Editor T. Asfour and Editor G. Venture upon evaluation of the reviewers’ comments. 
This work was supported by Institute of Information \& communications Technology Planning \& Evaluation (IITP) grant funded by the Korea government (MSIT) (No. RS-2019-II190079, Artificial Intelligence Graduate School Program (Korea University); No. 2022-0-00871, Development of AI Autonomy and Knowledge Enhancement for AI Agent Collaboration; and No. RS-2024-00336738, Development of Complex Task Planning Technologies for Autonomous Agent). 
% We thank NAVER LABS crop. for helping set up the AMBIDEX hardware for our experiment.
} %Use only for final RAL version
\thanks{$^{1}$Jeongeun Park, Taemoon Jeong, Hyeonseong Kim, Taehyun Byun, Seungyoun Shin, and Sungjoon Choi are with Department of Artificial Intelligence,
        Korea University, Seoul, Republic of Korea
        {\tt\small \{baro0906, taemoon-jeong, hyeonseong-kim, taehyun-byun, 2022021568, sungjoon-choi\}@korea.ac.kr}}%
\thanks{$^{2}$Keunjun Choi is with the Rainbow Robotics,
        Daejeon, Republic of Korea
        {\tt\small keunjun.choi@rainbow-robotics.com }. This work was conducted while at NAVER LABS.}%
\thanks{$^{3}$Jaewoon Kwon and Taeyoon Lee are with the NAVER LABS, Seongnam, Republic of Korea
        {\tt\small \{jaewoon.kwon, ty-lee\}@naverlabs.com}}%
\thanks{$^{4}$ Matthew Pan is with 
the Ingenuity Labs Research Institute and Department of Electrical and Computer Engineering, Queens University, Kingston, Canada {\tt\small matthew.pan@queensu.ca }}%
\thanks{Project page: \url{https://mask-robot.github.io.}}\label{page}
\thanks{Digital Object Identifier (DOI): see top of this page.}
}

% The paper headers
%\markboth{Journal of \LaTeX\ Class Files,~Vol.~14, No.~8, August~2015}%
%{Shell \MakeLowercase{\textit{et al.}}: Bare Demo of IEEEtran.cls for IEEE Journals}
\markboth{IEEE Robotics and Automation Letters. Preprint Version. Accepted July, 2024}
{Park \MakeLowercase{\textit{et al.}}: Masquerading Animated Social
Kinematic (MASK)}

% make the title area
\maketitle

% As a general rule, do not put math, special symbols or citations
% in the abstract or keywords.
\begin{abstract}
This paper presents the design and development of an innovative interactive robotic system to enhance audience engagement using character-like personas. Built upon the foundations of persona-driven dialog agents, this work extends the agent's application to the physical realm, employing robots to provide a more captivating and interactive experience. The proposed system, named the Masquerading Animated Social Kinematic (\method), leverages an anthropomorphic robot which interacts with guests using non-verbal interactions, including facial expressions and gestures. A behavior generation system based upon a finite-state machine structure effectively conditions robotic behavior to convey distinct personas.
The \method~framework integrates a perception engine, a behavior selection engine, and a comprehensive action library to enable real-time, dynamic interactions with minimal human intervention in behavior design. Throughout the user subject studies, we examined whether the users could recognize the intended character in both personality- and film-character-based persona conditions. We conclude by discussing the role of personas in interactive agents and the factors to consider for creating an engaging user experience.
\end{abstract}

% Note that keywords are not normally used for peerreview papers.
% \begin{IEEEkeywords}
% IEEE, IEEEtran, journal, \LaTeX, paper, template.
% \end{IEEEkeywords}
\begin{IEEEkeywords}
Social HRI; Gesture, Posture and Facial Expressions; Design and Human Factors
\end{IEEEkeywords}

\section{Introduction}
Interactivity in robots~\cite{20_pan,22_van,19_Mohammadi} can establish meaningful connections with humans, thereby greatly improving the user experience by creating engaging relationships. 
Integrating character-like attributes into these robots makes interactions more human-like and relatable, enhancing emotional connections and creating memorable experiences. 
Interactive robots embedded with customizable personas can offer companionship for combating loneliness or deliver standout performances in the entertainment sector. 
Leveraging this potential, we focus on building an interactive robotic system that takes a step towards adapting flexible personas that can take on the likeness of personalities or film characters. 

Building character-like or persona-driven dialog agents~\cite{23_shao,19_Mohammadi} have been actively studied to provide users with agents that experience events, express emotions, and interact with people. 
We believe interactive robotic agents could bring such agents into the real, physical world. Combining physical movements - i.e., gaze and gestures - with persona-driven agents can amplify the richness of human-machine interactions.

We hypothesize that by embedding a character template into robotic behavior, interactive agents can convincingly embody distinct personas for users to engage with. In particular, we aim to achieve this through a simple finite-state machine framework, which determines robotic actions from an observed user's behavior, a current robot's behavior, and a persona.
We focus on non-verbal communication, where the social cues come from gestures, gazes, or facial expressions. For instance, if a robot demonstrated actions representing fear when the user is greeting it, the users would be led to associate a \textit{``shy"} or \textit{``introverted"} character with the robot. Furthermore, our system aims to easily adopt diverse personas, similar to a robot putting on and switching masks, requiring minimal human effort to design such robots.

The system comprises the integration of three key components~\cite{20_pan}: a perception engine that extracts meaningful features from users' 3D body pose, a behavior selection engine that selects appropriate robotic behavior within the context, and an action library housing a collection of robot motions and facial expressions. We introduce a persona-infuser module to automate the process of generating behavior transitions with non-verbal cues and a given persona leveraging large language models (LLMs)~\cite{23_openai}, which is utilized in the behavior selection engine. 
Notably, our framework enables minimal human intervention in behavior selection design, enabling the swift creation of interactive robotic agents imbued with distinct personas through textual input. Additionally, we enable functionality to imitate well-known movie characters where robots can harness familiarity of their personas to allow users to quickly identify the characters being imitated, significantly enhancing user engagement and acceptance.

To test our system, we recruited 162 participants to analyze a persona-driven robot showcased in a public cafe. Results show the participants could recognize the robot's given persona in both personality and film character-based persona agents with captivating experiences. Based on participants' post-study comments, we identified opportunities and challenges in designing persona-based interactive robots.

\section{Related Work}

% persona-driven HRI
The influence of an agent's persona on a user's experience has been actively explored in various domains, including robotics applications \cite{18_mota} and dialog agents \cite{20_li, 23_safdari,23_lee}. Several papers \cite{18_mota,22_van,19_Mohammadi,23_Moujahid} have focused on building personality-driven agents to provide the user's engaging experience. Van Otterdijk et al.~\cite{22_van} focused on developing robot personalities expressed through nonverbal cues with distinct personality traits like extroversion and introversion.
Mohammadi et al.~\cite{19_Mohammadi} investigated whether a robot with a socially engaged personality is more accepted than one with a competitive personality through a case study involving a dice game. In addition, Moujahid et al.~\cite{23_Moujahid} introduced Charlie, a stationary robot receptionist designed to interact verbally and non-verbally in dynamic environments, focusing on how different robot personalities, specifically introversion and extroversion, affect human behaviors. However, previous work on robotic personas has often been restricted by the need for manual design of personalities by experts. In contrast, this work utilizes large language models (LLMs) to automate the adaptation of various personas, including personality and fictional characters. This approach reduces human effort and enhances the flexibility of persona development, particularly in automating the process of creating behavior models through finite-state machines.

% LLMs personas (Any language model)
Prior studies~\cite{20_li, 23_safdari,23_lee} have explored the ability of language models to adapt specific personas or personalities. 
% Li et al.~\cite{20_li} introduced Human Level Attributes (HLAs) based on tropes to create dialogue agents that can mimic the personalities of fictional characters. 
Safdari et al.~\cite{23_safdari} discussed the impact of synthetic personality in large language models (LLMs) on conversational agents, introducing a method for assessing and validating personality tests on LLMs and for influencing the personality in LLM-generated text. Lee et al.~\cite{23_lee} demonstrated distinct LLM-based behaviors that enable more authentic and context-aware human-robot interactions by integrating non-verbal cues. However, the approach has limitations in the latency of requesting API calls, and the explored persona was dialog-centric rather than the character or style of the robot. 

% non-verbal communication
Non-verbal cues play a significant role in conveying emotions and intents in communications. Non-verbal communication can be composed of gaze, gestures, or intonation. Pan et al.~\cite{20_pan} introduced a system that combines advanced gaze interaction technology and character animation principles with human-like gaze behaviors. Ko et al.~\cite{20_ko} focused on the motion of the robot to learn and generate the social behavior given the human pose. Brock et al.~\cite{20_brock} presented a real-time hand gesture recognition system to facilitate close-distance non-verbal communication with the tabletop robot Haru.
Aligning with the mentioned importance of non-verbal communication in HRI, we aim to build a system incorporating personas via non-verbal communication.

In our setup, visitors can interact with the robot non-verbally through body language. 
We hypothesize that interactive agents can convincingly embody distinct personas by selecting the next behavior appropriate to the user's non-verbal cues and to the robot's assigned personas (i.e., personalities or fictional characters). For instance, users could recognize the robot as \textit{``uncooperative"} persona after observing the robot rejecting the user's greeting.

\section{Proposed Method}

\begin{figure*}
    \centering
    \includegraphics[width=0.9\textwidth]{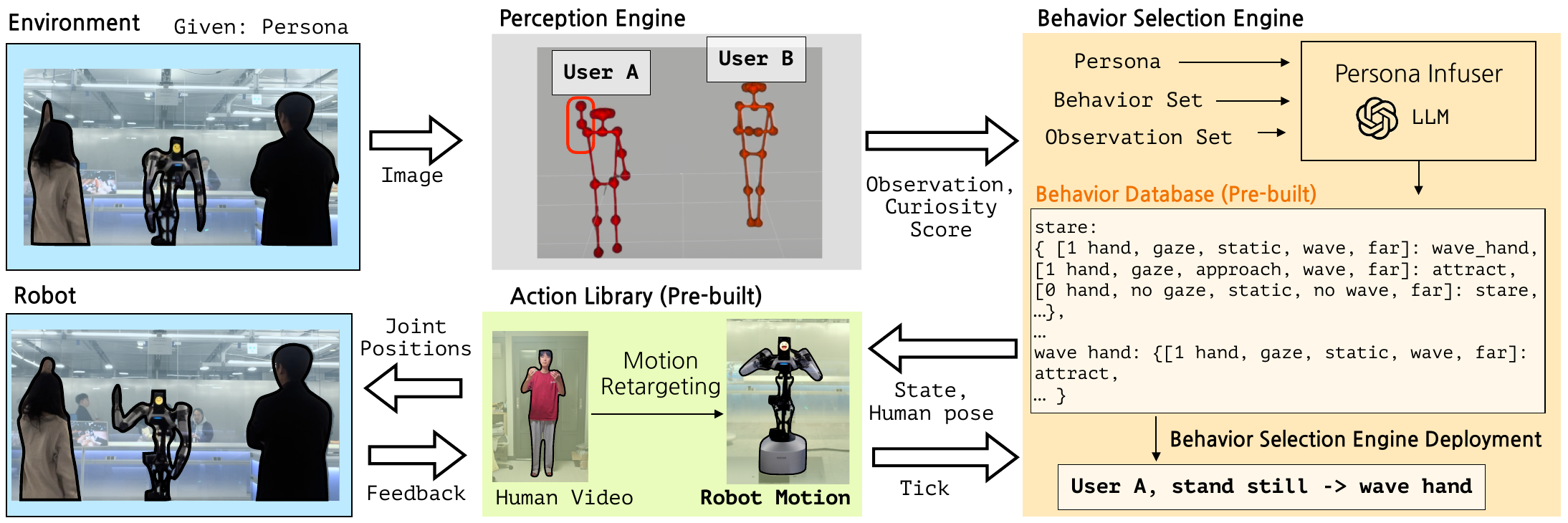}
    \caption{The proposed system architecture. The system is composed of a perception engine, a behavior selection engine, and an action library, where the action library and the behavior databases are pre-built components.}
    \label{fig:arc}
\end{figure*}

This section describes technical details for automating the design process in a persona-based interactive system. The system's architecture, as depicted in Figure \ref{fig:arc}, contains three main modules: a perception engine, an action library, and a behavior selection engine. At runtime, the perception engine estimates the 3D body poses of users seen in the environment via the Zed2i camera and refines human pose information to discrete observations and is used to obtain a ``curiosity score". 
% Details of how this score is calculated are described below. 
This data is used by the behavior selection engine, which contains the state machine defining the robot's persona-infused behavior. In the behavior selection engine, the robot selects the subsequent motion and the corresponding facial expression based on the observation and the current state. 
The engine's behavior database maps observations and the robot's persona to specific state-action transitions.
% A behavior database with this engine contains state-action transitions conditioned on the observations and the robot's persona. 
This behavior database is pre-built and populated by a persona-infuser, which leverages large language models~\cite{23_openai} to construct these transitions autonomously, alleviating the need for manual design. Finally, with the selected behavior, the robot physically displays the motion and the facial expressions that are stored within the action library. 

%combines pre-built elements such as the behavior library and the behavior database with real-time components. The behavior database contains state-action transitions conditioned on the observations and the robot's persona. This behavior database is populated by persona-infuser, which leverages large language models~\cite{23_openai} to construct these transitions autonomously, alleviating the need for manual design.In the behavior selection engine, the robot selects the subsequent motion and the corresponding facial expression based on the observation, the current state, and the behavior selection template. Finally, with the selected behavior, the robot physically displays the motion and the facial expressions from the behavior library. 
 
\subsection{Non-Verbal Cues}
\begin{table}[!t]
    \centering
    \begin{tabular}{|c||c|c|}
    \hline
    \multirow{5}{*}{Human} & \multirow{5}{*}{Observation (72)} & \# Raised Hands $\in \{0,1,2\}$,  \\
    & & Distance $\in$ \{close,far\}\\
    & & Gaze $\in$ \{gaze, no gaze\}\\
    & & hand velocity $\in$ \{waving, not waving\} \\
    & & approaching $\in$ \{approach, static, leave\}\\
    \hline
    \multirow{9}{*}{Robot}
    & \multirow{6}{*}{Motion (13)} & 1. wave hand big 2. wave hand small \\
    & & 3. look around  4. attract to come closer  \\
    & & 5. small bow 6. cry 7. push away \\
    & & 8. hide away 9. read book 10. standstill \\
    & & 10. yawn 11. teasing 13. cross arms \\
    % & & 12. standstill 13. cross arms\\
    \cline{2-3}
    & \multirow{4}{*}{\shortstack{Facial \\Expression (12)}} & 1. neutral  2. smile 3. cry  \\
    & & 4. angry 5. scared  6. excited   \\
    & & 7. reading book 8. confused 9. yes \\
    & & 10. tongue out 11. yawn 12. nod head \\
    \hline
    \end{tabular}
    \caption{Non Verbal Cues. The cues for human observation include four factors with 72 possible combinations, while robot cues contain 13 motions and 12 facial expressions with 156 possible combinations.}
    \label{tab:cues}
\end{table}
% persona
\method~employs non-verbal communication for human-robot interaction, with users interacting with the robot through body poses ($p$). As such, we examine non-verbal cues generated by non-verbal cues  (presented in Table~\ref{tab:cues}) to drive the interaction. 
Human non-verbal cues (observations $o(p)$) are defined as elements; e.g., number of raised hands, distance between human and robot, is human gazing at the robot, and hand velocity. 
These elements form the observation space ($\mathcal{O}$), encompassing all possible combinations of each observation element.
Robots' reactions utilize a discrete set of generated motions and facial expressions, which is the state space ($\mathcal{S}$), also presented in Table~\ref{tab:cues}. The state compromises the 156 possible combinations of 13 different motions and 12 different facial expressions. 
% Robot responses leverage motions and facial expressions, defined in a discrete set, which forms the state space ($\mathcal{S}$). 
% % These cues adhere to an established emotional model \cite{}, encompassing eight distinct emotional states. 
% Two arms, the waist, and a screen-based face enable the robot to express these cues, forming the state space ($\mathcal{S}$) consisting of all possible robotic motion and facial expression combinations.

% for robots

\subsection{Automated Persona Infuser via LLMs}

\begin{figure}[!t]
    \centering
\includegraphics[width=0.95\columnwidth]{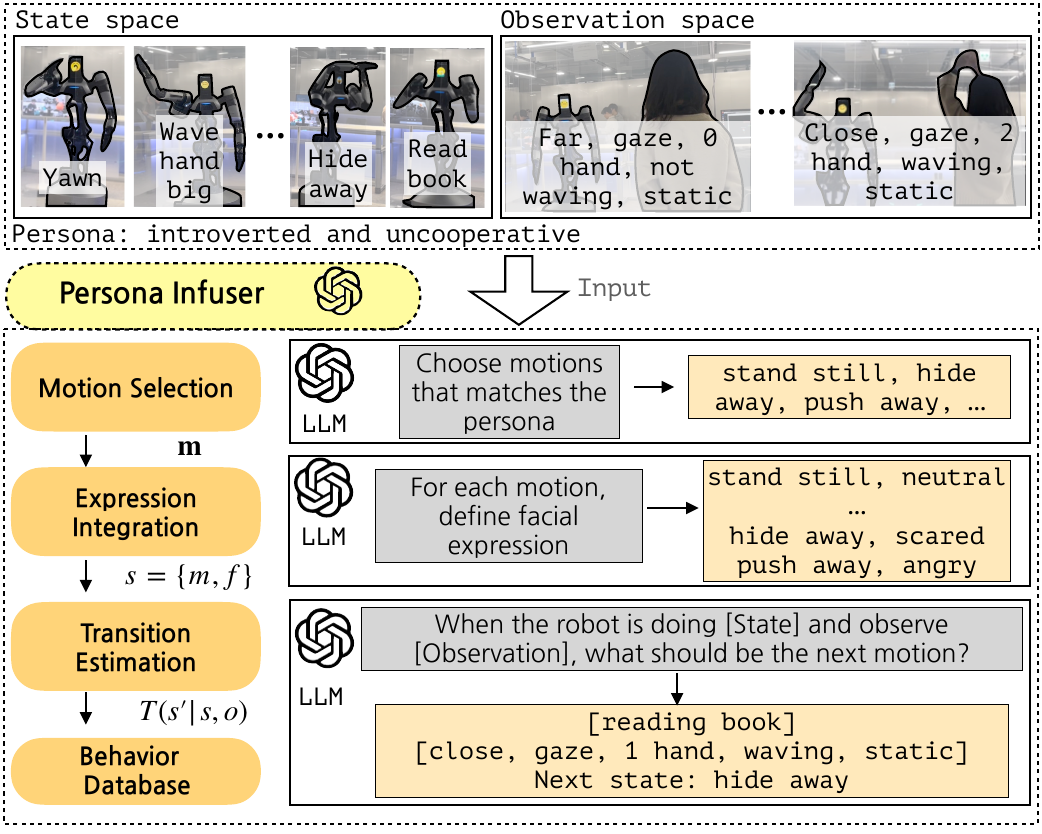}
    \caption{Illustration of the proposed Persona Infuser, which generates the behavior database via LLM. }
    \label{fig:llm}
\end{figure}
% intro
In the quest to embed robots with distinct personas, we introduce a persona infuser that constructs a behavior database. This database, $\mathcal{D}$, acts as a blueprint for persona-driven behavior, encoding all possible combinations of states, observations, and transitions by large language models~\cite{23_openai} (LLMs). The database is pre-built prior to runtime to reduce potential latency from language model inference, ensuring smooth and responsive robot actions. We utilize \texttt{gpt-4-0613} model\footnote{https://platform.openai.com/docs/models} for our \method~ system and our experiments. 
The input of the LLM~\cite{23_openai} becomes the defined state space $\mathcal{S}$, observation space $\mathcal{O}$, and the persona $x_p$. Conditioned with the mentioned inputs, the LLM estimates the state transitions $\mathcal{D} = \{T(s'|s, o)\}^{s \in \mathcal{S}, o \in \mathcal{O}}$. The behavior database is formulated as a dictionary. We design a hierarchical structure for generating behavior transitions as illustrated in Figure \ref{fig:llm}. 
% The behavior database is formulated as a dictionary type, illustrated as follows:

% \greybox{ \highlight{current state 1}:\\
%     \{
%         \observation{obs1}: \gray{next state 1} ,
%         \observation{obs2}: \gray{next state 2} ,
%           ... ,
%         \observation{obs N}: \gray{next state N} 
%     \}, 
%     ... , \\ 
%     \highlight{current state N}:\\
%     \{
%         \observation{obs1}: \gray{next state 1},
%         \observation{obs2}: \gray{next state 2},
%          ... , 
%         \observation{obs N}: \gray{next state N} 
%     \}
% }
% The behavior database $\mathcal{D}$ is defined as follows:
% \begin{equation}
    
% \end{equation}
% where $T$ is a transition to the next state $s'$. 

% hierarchical structure
% We design a hierarchical structure for generating behavior transitions. 
% Although LLMs have strength in reasoning abilities~\cite{22_ouyang,23_openai}, one of the drawbacks of LLM comes from a long sequence~\cite{23_liu}. 
% As handling a long sequence can cause forgetting issues in a complex environment, we adapt the hierarchical structure to build the behavior database. 
% We first select the motions that fit the given persona, then predict the transition given every state and the observation. Finally, we estimate the facial expressions corresponding to the motions for consistency in the gestures. The framework is illustrated in the Figure \ref{fig:llm}

\paragraph{Motion Selection}
We first utilize LLMs to predict the set of motions relevant to a given persona. This procedure helps the system filter out the states unrelated to the persona. The input of the LLM, denoted as $p_{\theta}$, includes the robot's motion cues $\mathcal{M}$, the target persona $x_p$, and the motion selection prompt $x_{\text{motion}}$. In the motion selection prompt, we enable the LLM to select between $0.25M$ and $0.5M$ different motions, where $M=13$ is the number of motion cues. 
% We believe that the knowledge of LLM can enable the system to select the relevant motions flexibly depending on the given persona. 
The predicted set of motions $\mathbf{m}$ is defined as 
\begin{equation}
    \mathbf{m} \sim p_{\theta} (\mathbf{m}|\mathcal{M}, x_p, x_{\text{motion}}).
\end{equation}

% The illustration of a motion selection prompt is as follows:\\
% \greybox{\sysprompt{You are} \highlight{[persona]} \contpropmt{robot that is interacting with users.} 
% \prompt{Choose at least ($0.25M$) and at most ($0.5M$) motions that you want to play, considering your persona. Please choose from the following motions:}
% \highlight{[motions]} \contpropmt{Try to choose different meanings of motions that can be used in different situations.}\\
% \LLM{wave\_hand\_big, look\_around, ...}
% }

\paragraph{Expression Integration}
In this step, we establish state $s$ that pairs each selected motion ($m$) with a corresponding facial expression ($f$). 
% In the second step, we aim to define a state $s$ composed of motion ($m$) and the corresponding facial expression ($f)$. 
LLM estimates the matching facial expression conditioned on motion, iterating the estimation through the selected motions $\mathbf{m}$. The state $s$ is defined as a pair of faces and the motion, and is as follows: 
$s = (f,m) \quad \text{where} \quad 
    f \sim p_{\theta}(f|m, \mathcal{F}, x_p, x_{\text{states}})$, 
% follows: 
% \begin{equation}
%     % s = (f,m) \quad \text{where} \quad 
%     % f \sim p_{\theta}(f|m, \mathcal{F}, x_p, x_{\text{states}})
% \end{equation}
$m \in \mathbf{m}$, $x_{\text{states}}$ is a prompt for expression integration, and $\mathcal{F}$ is a pre-defined set of facial expression cues. The set of states is denoted as $\mathbf{s} = \{(f,m)\}^{m \in \mathbf{m}}$. 
% The illustration of a state definition prompt is as follows:\\
% \greybox{
% \sysprompt{I am} \highlight{[persona]} \contpropmt{robot that is interacting with users. }
% \prompt{For each state, we want to estimate the facial expression of the motion. The facial expression should be in the following labels: }\highlight{[facial expressions]} \contpropmt{What would be the facial expressions for state} \observation{current motion} 
% \LLM{facial expression}
% }
In addition, we utilize LLMs to define an initial state $s_0 \sim p_{\theta}(s_0|\mathbf{s},x_p,x_{\text{init}})$, where $x_{\text{init}}$ is a prompt for defining initial state. The textual inputs of the states are designed as the names of the motion to simplify the notations to LLM. 
% The illustration of a state initialization prompt is as follows:\\ 
% % start state $s_0$
% \greybox{\sysprompt{I am} \highlight{[persona]} \contpropmt{robot that is interacting with users. I have a set of motions that I can do and a set of observations that I can observe from the user. I can conduct the following actions: \highlight{[selected states]}}. \prompt{
% which state should the robot start with? Please adapt the robot persona,} \highlight{[Persona]} 
% \LLM{start state}}

\paragraph{Transition Estimation}
Based on the persona-based state sets $\mathbf{s}$, we estimate the state transitions $T(s'|s,o)$ via LLMs. 
% We aim to build a transition that covers all the defined states ($\mathbf{s}$) and observation ($\mathcal{O}$) space. 
% Instead of utilizing LLMs to build the transitions with a single inference, we iterate our process $|\mathbf{s}| \times |\mathcal{O}|$ times to infer the next state, given the current and the observation. 
To infer the robot's next state, we repeat our evaluation process for each combination of current states ($|\mathbf{s}|$) and observations ($|\mathcal{O}|$), resulting in numerous iterations.
We formulate the deterministic state transition, where LLMs directly estimate the next state $s'$ given current state $s$ and observation $o$. The input of LLM is a state set $\mathbf{s}$ from the previous paragraph, the target persona $x_p$, current observation $o$, current state $s$, and the state transition prompt $x_{\text{transition}}$. The transitions are defined as follows:
\begin{align}
    T(s' |s,o) = \begin{cases}
        1 &  \text{if} \quad s' = s'_{\text{LLM}} \\
        0 & \text{otherwise}
    \end{cases} \\
    s'_{\text{LLM}} \sim p_{\theta}(s'|s, o,\mathbf{s},x_p, x_{\text{transition}})
\end{align}
where $s\in \mathbf{s}$ and $o \in \mathcal{O}$. 
Again, we iterate this process to cover all observation and state spaces.
% The illustration of a state transition prompt is as follows:\\
% \greybox{\sysprompt{I am} \highlight{[persona]} \contpropmt{robot that is interacting with users. I can interact with the users by motions, depending on the user's actions. I want to make a finite-state machine to interact with the user. I can conduct the following actions:} \highlight{[selected states]} \contpropmt{I have four different observations: hand\_up, distance, gaze, hand\_velocity}
% \prompt{what would be the next state from} \observation{current state} \contpropmt{?
% The robot observations:} \observation{observation}\contpropmt{
% Try to answer in a way that fits the persona of the robot and the current state.
% }
% \LLM{next state}
% }

% \paragraph{Stochastic}
% \blue{TODO}

\subsection{Perception Engine}
During the real-time deployment phase, the perception engine detects the user's body pose to determine the user's state. The input of the perception engine is an RGBD image, and the output is the observation and a curiosity score (a score that describes how `interesting' a person is based on observed kinematic quantities as used in \cite{20_pan}) for each person. The camera stays stationary during the interaction, and the 3D body pose of the users with tracking is estimated via ZED SDK 4.0\footnote{https://www.stereolabs.com/docs/body-tracking/}. 
% The skeleton data is transformed into the base of the robot frame. 
% We compute the curiosity score based on these skeleton data and discretize the observation based on the features. 
% attention score

% \paragraph{Observation Space}\label{sec:obs}
For each individual, we transform body skeleton data into five observation categories: the number of hands raised above nose level, eye gaze (based on a gaze function $g(p)$ threshold $t_g$), distance (`close' or `far' relative to threshold $t_d$), hand movement speed (`waving' or `not waving' determined by a hand velocity threshold $t_v$), and approaching velocity ($\hat{v}^{L_v}_{\text{noses}}$ classified as `approaching,' `leaving,' or `static' based on threshold $t_a$). 
The gaze function $g$ is defined as the cosine of the angle between the normal vector to the plane formed by the nose and both eyes (left and right) and the position vector of the nose, effectively providing a measure of the direction in which the nose is pointing relative to the plane of the eyes. For the approaching velocity, we track the $L^v$ second of the nose pose to determine whether the user is getting closer to the robot. 
These observations are summarized for each person as $o(p)$ in Table \ref{tab:cues}.

% \paragraph{Curiosity Score}
The curiosity score determines which user to interact with in multi-person scenarios, indicating how interested each user is to the robot. Following previous work~\cite{20_pan}, the curiosity score $\Phi$ for each person is defined as follows: 
\begin{equation}
\label{eq:cur}
\begin{split}
    \Phi(p) = \Theta(t) \cdot \big( w_d \|p_{\text{nose}}\| + w_h(h_{\text{right hand}}+ h_{\text{left hand}}) + \\w_v(\hat{v}_{\text{right hand}}+\hat{v}_{\text{left hand}}) + w_g g(p_{\text{eyes}}) + w_a \hat{v}^{L_v}_{\text{noses}}\big)
\end{split}
\end{equation}
where $p$ is a body pose, $w_d, w_v, w_h, w_g$ is a weight for distance, hand velocity, hand raise, and gaze, respectively. The curiosity score is composed of four elements: a distance from the robot $\|p_{\text{nose}}\|$, an indicator that determines raising hand $h_{\text{right hand}}, h_{\text{left hand}}$, a hand velocity $\hat{v}_{\text{right hand}}, \hat{v}_{\text{left hand}}$, a gaze parameter $p_{\text{eyes}}$, and an approaching velocity $\hat{v}^L_{\text{nose}}$. 
$\Theta$ is a habituation factor at the time $t$, which penalizes curiosity score for short observations. As mentioned in the previous paragraph, $g$ and $L_v$ are the gaze functions and approaching velocity. 

\subsection{Behavior Selection Engine}
The behavior selection engine selects the next states situated within the context. Based on observations, the robot's current state, and the behavior database, the selection engine selects the next state and the user to interact with. The user to interact with is determined via curiosity score. As
% We first formulate the guest base, where we save the users' information, which includes the users' positions, observations, curiosity scores, and the binary status of moving or static (the robot only attempts to interact with people if they are static). 
% In addition, the curiosity level is also tracked to determine the person to interact with. 
% refine the curiosity
% As 
curiosity in equation \ref{eq:cur} represents the curiosity from the current user's motion; we introduce a refined model for natural turn-taking. This model dynamically adjusts curiosity levels based on interaction time, decreasing curiosity for individuals who have already engaged with the robot and increasing it for those who have not. 
The refined curiosity $\Phi^{r}(p)$ is defined as follows: 
\begin{equation}
    \Phi^{r}(p) = \min\big(\begin{cases} 
        \Phi(p) / (\epsilon_d(\Phi(p)))^{t_{in}} &\text{where} \quad p = p_{in}\\
        \Phi(p) \cdot (\epsilon_i)^{t_{nn}} & \text{otherwise}
    \end{cases}, \phi_{max} \big)
\end{equation}
where $t_{in}$ is a time of interaction with the user for person $p$, $t_{nn}$ is a time of non-interaction, and $p_{in}$ is a person interacting with a robot. $\epsilon_d(\Phi(p))>1$ and $\epsilon_i > 1$ are the decrease rate and increase rate of the curiosity score, and $\phi_{max}$ is a maximum value for the curiosity. We design the curiosity decrease rate $\epsilon_d(\Phi(p))$ to be dynamic; the rate slows down for users expressing strong interest, allowing for extended interaction. 
% The decreasing rate is defined as follows: 
% \begin{equation}
%     \epsilon_d(\Phi(p))) = \begin{cases}
%         \epsilon_s & \text{where} \quad \Phi(p) < \phi_s\\
%         \epsilon_l & \text{otherwise}
%     \end{cases}
% \end{equation}
% where $\epsilon_s<\epsilon_l$, decrease rate depending on the curiosity level, and $\phi_s$ is threshold. If the person is not detected for $0.5$ seconds, the person is removed from the guest base. 
% \paragraph{behavior Selection}
% trigger
% Driven by the guest base and the pre-built behavior selection database, the system identifies the individual to interact with and selects the next state. 
Decisions for state change unfold at two key points: when changes occur in the observation or when a person of interest changes. 
The person of interest, which denotes a person that the robot is to interact with, is defined as a person with a maximum curiosity in the scene $p_{in} = \arg \max_{p} \Phi^{r}(\mathbf{p})$. Then, the next state $s'$ is predicted by the behavior database $T(s'|s,o(p_{in}))$, with the current state $s$ and the observation from the person of interest $o(p_{in})$. Finally, to face the user while interacting with them, we also obtain the person's heading direction $\theta_{p_{in}}$ with respect to the robot frame. The information about the next state and the heading direction is then transferred to the action library to display the robotic action. 

% from the state transition
% motion safety duration

\subsection{Action Library}

\begin{figure*}[!t]
    \centering
    \includegraphics[width=0.94\textwidth]{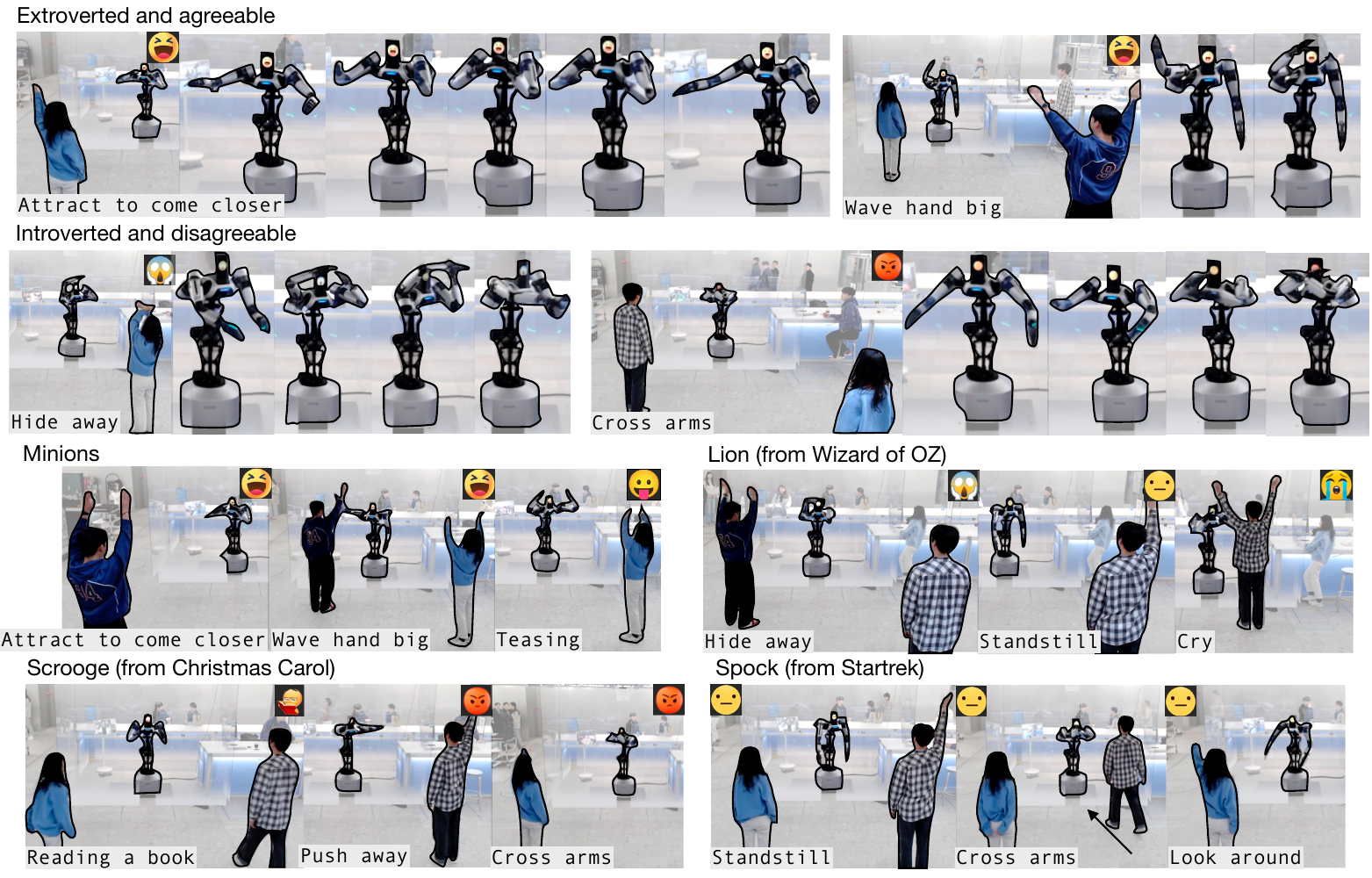}
    \caption{Demonstrations of the system. The first and second rows demonstrate the personality-based persona interaction, while the third and fourth rows demonstrate the character-based persona interaction. The emojis on the top indicate the facial expression that is displayed on the robot's screen. The caption under each image denotes the motion name.}
    \label{fig:demo}
\end{figure*}
The system is built upon a pre-built action library, where each state maps to specific joint trajectories and facial expressions. The action library module expresses the motion and facial expressions chosen from the behavior selection engine. Facial expressions are directly displayed on the screen as emojis.
% , and motions are presented in joint trajectories. 
% When designing the robotic motions for the interaction, we consider two main factors: motion expressiveness and transition smoothness.
% \paragraph{Motion Set}
The robot's motions are obtained from a robust motion retargeting pipeline that is described in~\cite{19_choi} where human-we recorded the demonstration of the desired motion suitable for human-robot interaction scenarios. 
% Each human motion is retargeted to the robot with smooth, expressive, and self-collision-free trajectories.
For the robots to face the user in the correct direction, we gave variants in the yaw direction at the waist. We have added yaw variants $\theta_w \in \{-\frac{\pi}{3} + \frac{\pi \cdot i}{9} \}_{i=0}^{6}$ at the waist joints to the original motion with the collision handling process for every motion. This process results in $13 \times 7 = 91$ motions in the library. We map the robot's yaw to the user's closest orientation during the runtime.
For the inbetweening trajectories for each motion, we utilize an optimization-based method aware of position and velocity limits. Each transition trajectory is defined to span from a second interval of every motion trajectory to the start of the next trajectory. 

% motion retargeting
% \paragraph{In-betweening Trajectories}
% We utilize an optimization-based method to create trajectories that are aware of position and velocity limits. As the robot shifts between different motions during its original motion, it needs to follow a smoothly interpolated trajectory. This trajectory accounts for the velocity at both the transition point and the beginning of the next motion. Each transition trajectory is defined to span from a second interval of every motion trajectory to the start of the next trajectory. 
% The transition trajectory $\mathbf{x}$ is obtained as follows: 
% % motion tweening
% \begin{equation}
% \label{eq:opt_2}
% \begin{split}
%     \text{minimize}\qquad & \frac{|| \mathbf{x} ||}{M} \\
%     \text{subject to} \qquad & \mathbf{x}^{min} < \mathbf{x} < \mathbf{x}^{max}\\
%     & |\ddot{ \mathbf{x}}| < a^{max},\quad |\dddot{\mathbf{x}}|< j^{max}\\
%     & \mathbf{x}_0 = \mathbf{x}_{init}, \quad \mathbf{x}_M = \mathbf{x}_{end} \\
%     & \dot{\mathbf{x}}_0 = v_{init}, \quad \dot{\mathbf{x}}_M = v_{end}
% \end{split} 
% \end{equation}

% where $M$ is the length of the trajectory, and the terms $\mathbf{x}^{min}, \mathbf{x}^{max}, a^{max}, j^{max}$ refer to the joint position, acceleration, and jerk limits, respectively. The equality constraints are the desired initial position ($\mathbf{x}_{init})$, final position ($\mathbf{x}_{end}$), initial velocity $(v_{init})$, and the final velocity ($v_{end}$). 

\subsection{Persona-Based Design Outcomes}

The snapshot of robotic behaviors based on various personas is illustrated in Figure \ref{fig:demo}. In the top row, the robot demonstrates an \textit{extroverted and agreeable} persona (based on personalities) through open and engaging body language, as indicated by the come hither beckoning and waving hands. In the second row, the robot adopts an \textit{introverted and disagreeable} persona, characterized by a less engaging body language, with arms crossed and hiding from the interaction partner. In the third and fourth rows, the robot embodies film characters—the playful yet mischievous \textit{Minions} and the \textit{Cowardly Lion from ``The Wizard of Oz,"} characterized by timid and fearful actions such as hiding or crying. The final rows feature the robot mirroring the personas of \textit{Scrooge from ``A Christmas Carol"} and \textit{Spock from ``Star Trek,"} where \textit{Scrooge} is represented with irate gestures, and Spock is depicted with keeping a neutral face to express an emotionless character. Additional visual materials, including video demonstrations and figures of the generated behavior database, are available on the project's \href{https://mask-robot.github.io}{webpage}.

\section{Experiments}
% \subsection{Hypotheses}
We conducted a human-subjects study to investigate the capabilities of the proposed method in representing the persona. 
Our main question is whether our system enabled users to distinguish robot's personas while considering the effects of persona-based interactions on user experience and engagement. 
Throughout the experiment, we aim to examine the hypothesis that the system will enable users to identify the robot's persona. 
% In particular, we set two hypotheses based on the type of the persona: (1) There will be a significant difference in the user's recognized personality between different robot personality types (Introverted / Extroverted, Agreeable / Disagreeable), (\textbf{H1}). 
% (2) People can successfully infer the intended character with character-based persona agents (\textbf{H2}). 
% \review{We define a $0.61$ F1-score as successfully classifying the correct persona. This standard is consistent with the highest performances observed in related research within the field, demonstrating its reliability and comparability to established methodologies \cite{9207529}.}
% We define \review{$0.61$ F1-score as successfully classifying the correct persona, mirroring the highest score achieved in related research \cite{9207529}.} 
% \begin{itemize}
%     \item \textbf{H1:} There is a significant difference in the user's recognized personality between different robot personality types.
%     (Introverted / Extroverted, Agreeable / Disagreeable)
%     % \item \textbf{H2:} The impact of facial expression is insignificant. 
%     \item \textbf{H2:} People can successfully infer the intended character with character-based persona agents. 
% \end{itemize}
\subsection{Experimental Setup}

The platform used in this work is the Ambidex~\cite{20_choi}, a tendon-driven mechanism consisting of a head, waist, and two arms. 
The robot features 32 degrees of freedom: 12 in the waist and 10 in each arm, which allows for enhanced expressiveness in non-verbal communication through body language. 
A screen that forms the head of the robot displays the robot's facial expressions. 
% The robot utilizes a 1 kHz low-level control system that computes the required torque for controlling the specified joint positions of the robot. This low-level controller has enhanced the bandwidth of cable-driven robots and improved tracking performance. 
An RBGD camera placed at the robot's top detects the user's body pose. The robot is encased within a transparent acrylic barrier to ensure interaction safety. 
For this work, the robot was placed in the cafe in NAVER 1784\footnote{https://www.navercorp.com/en/naver/1784}, a space open to the public.

The experiment was conducted in two phases.
% During the first phase of the survey, the participants were exposed to personality-based persona robots, and the character-based persona robot was showcased to the participants in the second phase. 
In both phases, participants were asked to freely interact with the robot for approximately a minute. Following this interaction, participants were asked to fill out a survey (delivered in Korean). In phase 1, we first hypothesize that there will be a significant difference in the user's recognized personality between different robot personality types (Introverted / Extroverted, Agreeable / Disagreeable), (\textbf{H1}). The robot was show-cased with a personality-based persona to examine this hypothesis. We have chosen extroversion and agreeableness dimensions from the Big Five personality traits~\cite{92_mccrae}, as those two dimensions in building a social network~\cite{10_selfhout}, a key factor in building captivating personas in the engagement scenario. To keep the experimental procedure as short as possible for each participant (as we were surveying people who were passing by), we opted to use 1-item scale for each personality trait that is rated on continuous scale from 0 to 100. 
% . Agreeableness and extroversion are powerful dimensions in building a social network~\cite{10_selfhout}, a key factor in building immersive personas in the engagement scenario. 
 Four different personas created from two levels of each factor were used: \textit{extroverted and agreeable, introverted and agreeable, extroverted and disagreeable, and introverted and disagreeable}. In prompting a large language model, the term ``cooperative" was specifically chosen over ``agreeable" for prompt engineering to model behavior more effectively in scenarios requiring acknowledgment and interaction. 
 A between-subjects study with $52$ participants (gender: M=26, F=26, ages: 21-57) was conducted where the participants were asked to answer a questionnaire after interacting with the robot having one of the persona.
 % The questionnaire asked participants to score the following on continuous scales from $0$ to $100$:
 % \textbf{S1:} Score the extroversion of the robot (Introverted - Extroverted), \textbf{S2:} Score the agreeableness of the robot (Disagreeable - Agreeable). 
 % , and \textbf{S3:} Overall Satisfaction of the interaction. 
 % All the questions were asked in continuous score from $0$ to $100$.
 \textbf{S1:} Score the extroversion of the robot (Introverted - Extroverted) 
 \textbf{S2:} Score the agreeableness of the robot (Disagreeable - Agreeable) \textbf{S3:} What is your overall satisfaction of the interaction. 
  % \begin{itemize}
    % \item \textbf{S1:} Score the extroversion of the robot (Introverted - Extroverted)
    % \item \textbf{S2:} Score the agreeableness of the robot (Disagreeable - Agreeable)
    % \item \textbf{S3:} What is your overall satisfaction of the interaction
 % \end{itemize}
 % We conducted pre-studies without using facial expressions with 16 participants and failed to find a significant difference attributable to the presence or absence of facial expressions in persona identification. 
 % As the emojis were displayed on the screen, we had to check whether facial expressions played a significant role in identifying the persona than body motion. The study did .
% To test the hypothesis \textbf{H2}, we have recruited an additional $16$ participants (gender: M=8, F=8, ages: 25-61) who were asked to fill in the same questionnaire after interacting with a random persona-equipped robot that does not have a facial expression. 

 % \paragraph{Character Classification}
During the second phase of our study, we conducted observational studies on user perceptions as they interacted with the robot by measuring the classification accuracy of personas. The robot was endowed with four distinct personas based on fictional movie characters, including:
 % In the second phase of our experiment, We aimed to e where four different personas, based on famous fictional movie characters, were loaded onto the robot; these included: 
 \textit{Ebenezer Scrooge from A Christmas Carol~\cite{dickens1858christmas}, a Minion from the Despicable Me and Minions franchise~\cite{10_minion}, The Cowardly Lion from The Wizard of OZ~\cite{baum1900wonderful}, and Spock from the Star Trek franchise~\cite{79_spock}}. We recruited 108 participants (gender: M=47, F=63, ages: 20-53) to participate in this between-subjects study, where the personas were randomly assigned to participants with equal numbers. Participants were asked to interact with the robot and answer three questions in the following order: 
 \textbf{S4:} List keywords that describe the robot's personality or behavior.
\textbf{S5:} Which of the following characters does the robot's behavior seem to be most similar to?
(multiple choice: The Cowardly Lion from The Wizard of OZ, A Minion from Despicable Me, Ebenezer Scrooge from A Christmas Carol, Captain America from The Avengers, Sloth from Zootopia, Spock from Star Trek)
\textbf{S6:} Rate how well the robot imitates the personality of the character you selected in the previous question. 
% \textbf{S5:} Which of the following does the robot's behavior seem closer to? (multiple choice: Lions in Wizard of OZ, Minions, Scoorege from Christmas Carol, Captain America, Slot from Zootopia, Spock from Startreck), \textbf{S6:} How much does the robot fit the persona selected above? (Continuous Scale $0$ to $100$). 
 % \begin{itemize}
 %    \item \textbf{S4:} List keywords that describe the robot's personality or behavior.
 %    \item \textbf{S5:} Which of the following characters does the robot's behavior seem to be most similar to?
 %    (multiple choice:The Cowardly Lion from The Wizard of OZ, A Minion from Minions, Ebenezer Scrooge from A Christmas Carol, Captain America from The Avengers, Sloth from Zootopia, Spock from Star Trek)
 %    \item \textbf{S6:} Rate how well the robot imitates the personality of the character you selected in the previous quesiton (Continuous Scale $0$ to $100$). 
 % \end{itemize}
 We also provided a brief summary of each character's personality to help users understand the character mentioned in the choices\footnote{ This research was carried out with ethics approval from the ethics review board of Korea University under proposal KUIRB-2024-0069-01.}.

\begin{figure}
    \centering
    \includegraphics[width=0.49\textwidth]{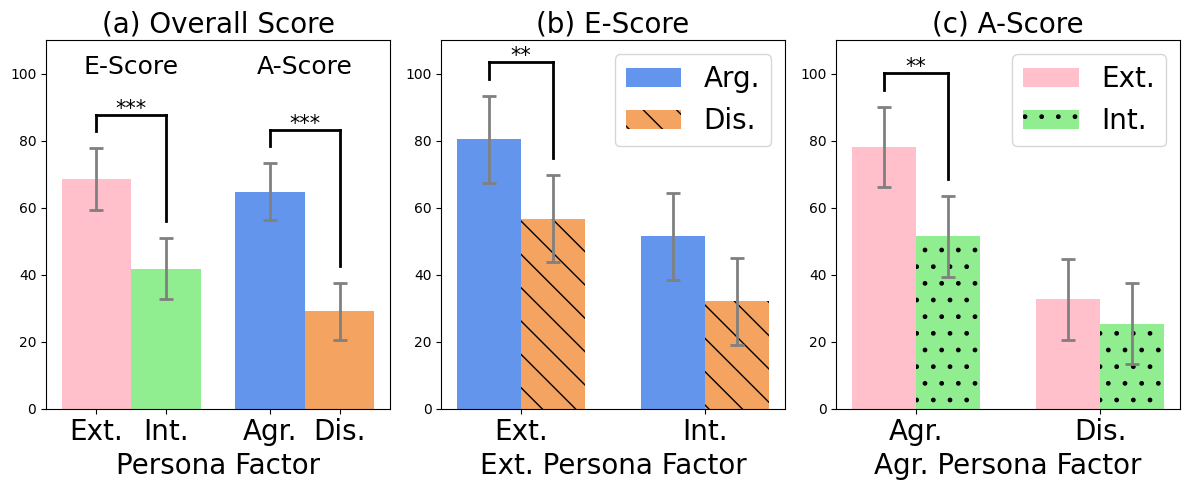}
    \caption{Results from survey phase 1 with questions S1 and S2. (a) E-score of extroverted factor and A-score of agreeable factor. (b) E-score of extroverted persona with agreeable persona. (c) A-score of agreeable persona with extroverted persona. Agr., Dis., Ext., and Int. denote agreeable, disagreeable, extroverted, and introverted personas, respectively. * as $0.01<p<0.05$, ** as $p<0.01$, and *** as $p<0.001$. The error bars represent 95\% confidence intervals. }
    \label{fig:res_persona}
\end{figure}

\subsection{Human Subject Study Results}
\paragraph{Personality Persona} \label{sec:pp}

% 2way MANOVA
% p values 
To test hypothesis 1 (\textbf{H1}), we conducted a two-way repeated-measures MANOVA on the results obtained from user survey questions \textbf{S1} and \textbf{S2} with $\alpha = 0.05$. These questions measure the dependent variables of extroversion and agreeableness scored by users for a specific displayed persona and are denoted here as the E-score and A-score, respectively. The persona is generated on the two different axes of personality we studied: extroversion and agreeableness. We observed that the users could recognize both extroversion and the agreeableness of the robot, as shown in Figure~\ref{fig:res_persona}-(a). The measured E-score shows a significant difference between extroverted and introverted persona ($F(1, 48) = 17.27$, $p < 0.001$, partial $\eta^2 = 0.265$). In addition, the persona based on agreeableness resulted in a significant difference in the A-score of the robot ($F(1, 48) = 35.40 $, $p < 0.001$, partial $\eta^2 = 0.424$).
No significant interaction effect was detected. We observed that the users feel extroverted robots are more agreeable and vice versa, as shown in Figure~\ref{fig:res_persona}-(b), (c). This resembles the previous findings from personality-based robots and psychology that extroversion and agreeableness are often positively correlated~\cite{21_andriella, 16_tov}. In analyzing the satisfaction of the interaction (\textbf{S3}), we conducted a two-way ANOVA and observed significant differences between agreeable and disagreeable personas ($F(1,48) = 4.799$, $p = 0.033$, partial $\eta^2 = 0.091$); users reported being more satisfied when interacting with agreeable personas. The facial expressions displayed by the robot helped users understand the context of the motions, but we failed to find a significant impact of using facial expressions during pilot studies.

\paragraph{Character Persona}\label{sec:sp}

\begin{figure}
    \centering
    \includegraphics[width=0.45\textwidth]{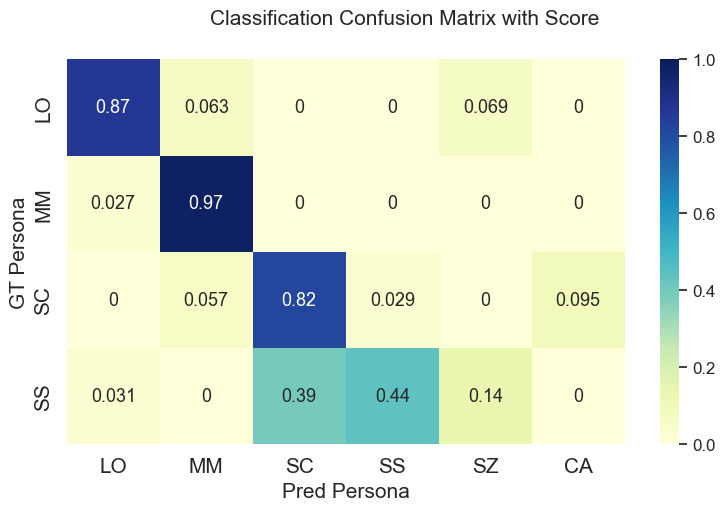}
    \caption{Classification Confusion Matrix for the character-based persona representing the characters \textit{Cowardly Lion (LO), Minion (MM), Scrooge (SC), Spock (SS), Sloth (SZ), and Captain America (CA), respectively.}}
    \label{fig:confusion}
\end{figure}
We measured the classification accuracy among characters based on user responses to survey questions \textbf{S5} and \textbf{S6} to examine how accurately users can identify each persona. We have observed that the participants could successfully identify the correct movie character with an average accuracy of $76.7\%$, $4.6$ times higher than random guess ($16.7\%$). 
% We examine hypothesis 2 (\textbf{H2}) using classification accuracy among characters based on the results of survey questions \textbf{S5} and \textbf{S6}. 
% The \review{average F1-score was reported as $0.78$, higher than the $0.61$ threshold set in\textbf{H2}}. 
The confusion matrix for classification is shown in Figure~\ref{fig:confusion}, where we have adjusted the user's fitting score obtained from the survey question \textbf{S6}. 
% \review{The confusion matrix is calculated by analyzing how often participants correctly identify a character, adjusted by their fitting scores from a question \textbf{S6}.}
The confusion matrix $T_{ij}$ is defined as 
\begin{equation}
    T_{ij} = \frac{\sum_{k} \big( s_f(k)\cdot\mathds{1}(c_k = c_i) \cdot\mathds{1}(g_k = c_j) \big)}{\sum_{k} \big( s_f(k) \cdot\mathds{1}(g_k = c_j) \big)}
\end{equation}
where $k$ is an index of a participants, $s_f(k)$ as a fitting score observed from \textbf{S6}, $c_k$ as a chosen character from \textbf{S5}, and $g_k$ as a persona shown to the participant $k$. 
We observe that users could successfully identify the \textit{Cowardly Lion from The Wizard of OZ, Minions,} and \textit{Scrooge} characters with great accuracy. However, participants appeared to misclassify the character of \textit{Spock} with \textit{Scrooge} ($0.39$). Two participants who interacted with the \textit{Spock} persona mentioned that the robot's firm and blunt behavior with a neutral face leads users to feel that the robot is \textit{``grumpy"}, making it easily confused with \textit{Scrooge}. 
% For keeping \textit{neutral} face in \textit{Spock} persona, 
% participant No. 68 commented that it played a crucial role in recognizing the character.
% while Participant No. 103 was confused about whether the robot was interacting with participants. was interactive?

\paragraph{Qualitative Result}
From \textbf{S4}, the users were asked to mention the keywords associated with each persona before answering \textbf{S5}. Again, the survey was conducted in Korean; the keywords are the closest translations available. Table \ref{tab:keywords} describes the top-three mentioned keywords for each character. Participants attributed traits matching each character's known qualities: ``shy" for the Cowardly Lion, ``playful" for the Minions, ``angry" for Scrooge, and ``indifferent" for Spock.
% For the persona of the \textit{Cowardly Lion,} participants most frequently used the terms ``shy," ``coward," and ``defensive," reflecting the character's well-known traits. In the case of the \textit{Minions} persona, the descriptors ``naughty," ``active," and ``playful" were highlighted, capturing the spirited and mischievous nature of the persona. \textit{Scrooge} robot was associated with negative attributes such as ``angry," ``aggressive," and ``dislike of others," which align with the character's portrayal in the classic story. Finally, \textit{Spock} agent was identified with the keywords ``pococurante," ``blunt," and ``cynical," capturing his less responsive and straightforward manner. 
These keywords from \textbf{S4} indicate that users accurately perceive the unique characteristics of persona-driven robots. 
% Table \ref{tab:keywords} describes the top-3 mentioned keywords for each character from survey question \textbf{S4} during the survey phase 2. \blue{TODO}

Throughout the detailed comments, we observe that most participants could understand the context of the behavior. One participant commented that \textit{``Judging by the constant peek-a-boo, [the robot] seemed like a friend with a strong sense of playfulness. Also, since it appeared to yawn when we didn't move, it seemed to get bored quickly if not engaged, representing a mischief character."} after interacting with the \textit{Minion} persona. 
More than half of the participants ($15$ out of $27$) who interacted with \textit{Minions} persona commented positively about active engagement and playful character.
We also observed that some participants created their own stories of the show, inducing a new interpretation of their behaviors. For the \textit{cowardly lion} persona, one participant commented that \textit{``The robot seemed to interpret the action of waving arms as an act of aggression, taking a defensive posture and shedding tears as if it was afraid of humans."}. However, another user reported that \textit{``It is impossible to understand the message it is trying to convey, and I am not feeling the interaction."} after interacting with \textit{Spock}, pointing out the lack of interpretability in unexpressive behaviors. 

\begin{table}[]
    \centering
    \begin{tabular}{|c|c|}
    \hline
        Persona & Keywords \\
        \hline
        Lion from Wizard of OZ & Shy, Coward, Defensive \\
        Minions & Naughty, Active, Playful\\
        Scrooge from Christmas Carol & Angry, Aggressive, Dislike others\\
        Spock from Startrek & Pococurante, Blunty, Cynical\\
        \hline
    \end{tabular}
    \caption{Top 3 mentioned keywords for each persona from survey question 4 (\textbf{S4}). }
    \label{tab:keywords}
\end{table}
% Table \ref{tab:keywords} presents a summary of the top three keywords that survey participants associated with each of the personas based on Survey question S4.

\section{Discussion}

% What are we going to discuss about! 

% summary
Throughout our user studies, we have shown that the participants could successfully recognize and distinguish the robot's persona in both personalities in Section \ref{sec:pp} (52 participants) and characters in Section \ref{sec:sp} (108 participants). 
In this section, we discuss the lessons learned during our studies and the main limitations of our work.

% Lessons learned: 
% 1. human's expectation regarding an interactive robot 
% 2. anthropomorphism
%
% Limitations
% - lack of enough actions
% - lack of enough observations
% - small number of states per persona -> leading to repeated motions
% - limitations of turn-taking (inability to interact with multiple people)
% - in character interaction, motions are not tailored to that character + appearance difference!! 
%

\paragraph{Human's Expectation Towards Robot}
We found discrepancies in user expectations versus the robot's behavior, as most participants were expecting warm and competent robot behaviors. We observed what appears to be a major user expectation mismatch regarding disagreeable personas, resulting in a notable decline in satisfaction levels. This trend indicates an expectation mismatch, as participants were unprepared for disagreeable personas. Furthermore, participants expected a competent robot that could respond to users rapidly and proactively.  
However, with inexpressive characters, such as the \textit{Spock} persona where the robot maintained a static state in response to users' greetings, participants found it challenging to understand the context of the interaction and sometimes believed the robot was malfunctioning.
This underlines the need for discreet selection of a robot's persona or additional cues to inform users in addition to non-verbal behaviors representing personas that consider the user expectations and the interaction context.

\paragraph{Anthropomorphism}
We noticed a tendency among participants to anthropomorphize the robot's actions as participants crafted their own narratives and interpretations of the robot's behaviors within the presented scenarios. 
For instance, one participant interpreted that the robot perceived the user's hand waving as a threat after observing defensive motions from the robot. 
Additionally, two participants noted that the robot with a playful character (\textit{Minion}) seemed to exhibit boredom when not greeted.
We believe the users had an engaging interaction by actively interpreting the robot's behavior and creating their own story in the show. 
This suggests that our research offers insights into how anthropomorphism influences user engagement and perception, highlighting the significant role of user interpretation in designing interactive robots. 
% Even though we have provided an explanation about each character, we carefully note that the character's familiarity may have affected the results, as 52\% of the participants were under age 30 and 94\% were under 40. 

% FSM limitations
% complexity of interaction 
% 200 participants
% 

% Limitations
% - lack of enough actions
% - lack of enough observations
% - small number of states per persona -> leading to repeated motions
% - limitations of turn-taking (inability to interact with multiple people)
% - in character interaction, motions are not tailored to that character + appearance difference!! 
%
\paragraph{Limitations}
Despite the proposed method's capability of mimicking characters or embodying personalities, our system currently suffers from some limitations and shortcomings. 
Most significantly, users reported a lack of diversity in robotic behavior for interactions that lasted beyond a minute. The proposed system does not offer a sufficient variety of actions and observations. Furthermore, the limited number of states available for each persona has resulted in users perceiving the robot as engaging in repetitive behavior.
This highlights a significant drawback to our system which should be considered in future iterations: added novelty and robotic behaviors for longer interactions will require a larger observation space to detect more non-verbal cues, and a library of robot actions that can be drawn by a more complex state machine within the behavior selection engine. Ultimately, a state machine architecture is not preferred as its complexity increases exponentially with added states - a new method of selecting appropriate behaviors would need to be implemented.

In addition, in group settings, we observed a delay in responding to the user's actions; when multiple people attempted to interact with the robot simultaneously, the robot could not interact with all of the users simultaneously. We have seen cases where users' curiosity scores decreased as the robot was still in states that responded to other participants. A more complex turn-taking system~\cite{23_paetzel} should be considered rather than a single curiosity score for enhancing the interaction quality. Finally, in character-based persona interaction, motions are not tailored to reflect the character, and there is a difference in the appearance of the character and the robot (e.g., AMBIDEX~\cite{20_choi} is too big to look like \textit{Minion}). 
These two factors contributed to participants noticing a gap between the intended character portrayal and the robot. Therefore, a review of the method for stylizing movements and potential design modifications should be taken into account.

% This is also related to the limitation of the structure, as expanding the finite-state machine structure increases the complexity of the system exponentially. Although our system automates the behavior selection process of finite-state machines, the kinematic transition still remains 

\section{Conclusion}
We have presented a system which we call \method~for interacting with humans non-verbally based on personas that can take on the likeness of personalities or film characters. This system integrates several components (perception engine with body pose estimation, behavior selection engine that incorporates an LLM for behavior selection, and action library with actions created through a motion retargeting pipeline) to take steps towards generating autonomous, interactive robot characters. As evidenced through user subject studies, we saw that conditioning robotic behavior with a state, an observation, and a persona can enable users to experience a convincing persona in interactive agents. We observed that users could recognize the intended character in both personality- and film-character-based personas. 
Despite the technical limitations of our system, such as lack of diversity in behavior and delay of response in group setting scenarios, we have shown preliminary steps towards bringing autonomous character-like agents into the physical world. 
We anticipate that this work will inspire future efforts toward expressive and convincing persona-based robotic agents that provide captivating experiences to people.
% \newpage
\bibliographystyle{unsrt}
\bibliography{reference}

\end{document}